\documentclass[10pt,twocolumn,letterpaper]{article}

\usepackage[pagenumbers]{cvpr} 

\usepackage{graphicx}
\usepackage{amsmath}
\usepackage{amssymb}
\usepackage{booktabs}

\usepackage{adjustbox}
\usepackage{multirow}
\usepackage{xcolor}

\usepackage[pagebackref,breaklinks,colorlinks,citecolor=teal]{hyperref}

\usepackage[capitalize]{cleveref}
\crefname{section}{Sec.}{Secs.}
\Crefname{section}{Section}{Sections}
\Crefname{table}{Table}{Tables}
\crefname{table}{Tab.}{Tabs.}

\def\cvprPaperID{791} 
\def\confName{CVPR}
\def\confYear{2023}

\begin{document}

\title{The Devil is in the Points: Weakly Semi-Supervised Instance Segmentation \\via Point-Guided Mask Representation}

\author{
Beomyoung Kim$^{1,2}$\hspace{1.5em}Joonhyun Jeong$^{1,2}$\hspace{1.5em}Dongyoon Han$^{3}$\hspace{1.5em}Sung Ju Hwang$^{2}$\\ \\
{NAVER Cloud, ImageVision$^1$\hspace{3em}KAIST$^2$\hspace{3em}NAVER AI Lab$^3$}\\
}

\maketitle

\begin{abstract}
    In this paper, we introduce a novel learning scheme named weakly semi-supervised instance segmentation (WSSIS) with point labels for budget-efficient and high-performance instance segmentation.
    Namely, we consider a dataset setting consisting of a few fully-labeled images and a lot of point-labeled images.
    Motivated by the main challenge of semi-supervised approaches mainly derives from the trade-off between false-negative and false-positive instance proposals, we propose a method for WSSIS that can effectively leverage the budget-friendly point labels as a powerful weak supervision source to resolve the challenge.
    Furthermore, to deal with the hard case where the amount of fully-labeled data is extremely limited, we propose a MaskRefineNet that refines noise in rough masks.
    We conduct extensive experiments on COCO and BDD100K datasets, and the proposed method achieves promising results comparable to those of the fully-supervised model, even with 50\% of the fully labeled COCO data (38.8\% vs. 39.7\%).
    Moreover, when using as little as 5\% of fully labeled COCO data, our method shows significantly superior performance over the state-of-the-art semi-supervised learning method (33.7\% vs. 24.9\%).
    The code is available at \url{https://github.com/clovaai/PointWSSIS}.
\end{abstract}
\section{Introduction}
\label{sec:intro}

Recently proposed instance segmentation methods~\cite{(MRCNN)he2017mask,(solov2)wang2020solov2,(mask_transfiner)ke2022mask,(SOLQ)dong2021solq,(queryinst)fang2021instances,(HTC)chen2019hybrid,(condinst)tian2020conditional,(blendmask)chen2020blendmask,(centermask)lee2020centermask,(yolact)bolya2019yolact} have achieved remarkable performance owing to the availability of abundant of segmentation labels for training.
However, compared to other label types ($e.g.,$ bounding box or point), segmentation labels necessitate delicate pixel-level annotations, demanding much more monetary cost and human effort. Consequently, weakly-supervised instance segmentation (WSIS) and semi-supervised instance segmentation (SSIS) approaches have gained attention to reduce annotation costs. WSIS approaches alternatively utilize inexpensive weak labels such as image-level labels~\cite{(BESTIE)kim2022beyond,(irn)ahn2019weakly,(prm)zhou2018weakly}, point labels~\cite{(BESTIE)kim2022beyond,cheng2022pointly,(wisenet)laradji2020proposal} or bounding box labels~\cite{(BBAM)lee2021bbam, (boxinst)tian2021boxinst,(tightness)hsu2019weakly}.
Besides, SSIS approaches~\cite{(NB)wang2022noisy,(shapeprop)zhou2020learning} employ a small amount of pixel-level (fully) labeled data and a massive amount of unlabeled data.
Although they have shown potential in budget-efficient instance segmentation, there still exists a large performance gap between theirs and the results of fully-supervised learning methods.

\begin{figure}[t]
    \centering
    \includegraphics[width=0.95\linewidth]{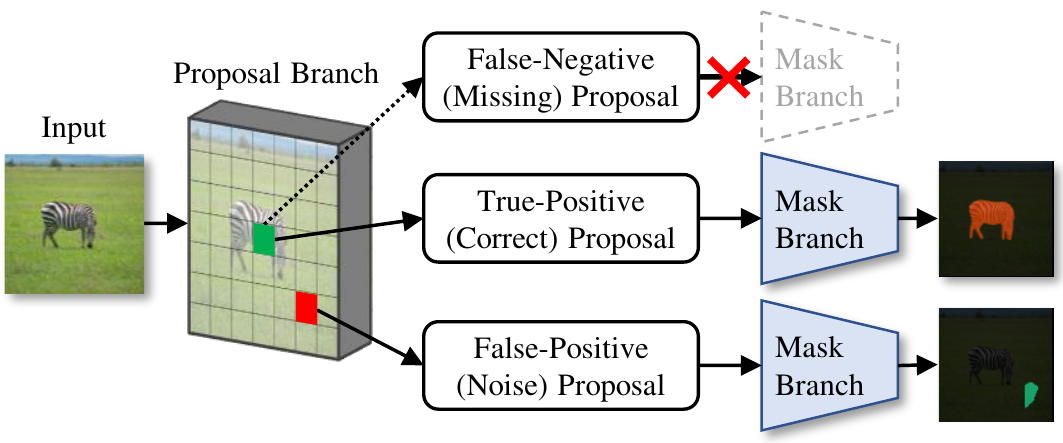}
    \caption{
        \textbf{Proposals and instance masks}. The absence of a proposal leads to the missing mask, even though the mask could be generated if given the correct proposal (zebra). Also, noise proposal often leads to noisy masks. Our motivation stems from the bottleneck in the proposal branch, and this paper shows economic point labels can be leveraged to resolve it.
    }
    \label{fig:motivation}
    \vspace{-2mm}
\end{figure}

Specifically, SSIS approaches often adopt the following training pipeline: (1) train a base network with fully labeled data, (2) generate pseudo instance masks for unlabeled images using the base network, and (3) train a target network using both full and pseudo labels.
The major challenge of SSIS approaches comes from the trade-off between the number of missing ($i.e.,$ false-negative) and noise ($i.e.,$ false-positive) samples in the pseudo labels.
Namely, some strategies for reducing false-negatives, which is equivalent to increasing true-positives, often end up increasing false-positives accordingly; an abundance of false-negatives or false-positives in pseudo labels impedes stable convergence of the target network.
However, optimally reducing false-negatives/positives while increasing true-positives is quite challenging and remains a significant challenge for SSIS.

To address this challenge, we first revisit the fundamental behavior of the instance segmentation framework.
Most existing instance segmentation methods adopt a two-step inference process as shown in Figure \ref{fig:motivation}: (1) generate instance proposals where an instance is represented as a box~\cite{(MRCNN)he2017mask, (HTC)chen2019hybrid, (centermask)lee2020centermask,(msrcnn)huang2019mask} or point~\cite{(solov2)wang2020solov2, (condinst)tian2020conditional,(solo)wang2020solo,(centermask)wang2020centermask} in proposal branch, and (2) produce instance masks for each instance proposal in mask branch.
As shown in Figure \ref{fig:motivation}, if the network fails to obtain an instance proposal ($i.e.,$ false-negative proposal), it cannot produce the corresponding instance mask.
Although the network could represent the instance mask in the mask branch, the absence of the proposal becomes the bottleneck for producing the instance mask.
From the behavior of the network, we suppose that addressing the bottleneck in the proposals is a shortcut to the success of the SSIS.

Motivated by the above observations, we rethink the potential of using point labels as weak supervision. 
The point label contains only a one-pixel categorical instance cue but is budget-friendly as it is as easy as providing image-level labels by human annotators~\cite{(what_point)bearman2016s}.
We note that the point label can be leveraged as an effective source to (i) resolve the performance bottleneck of the instance segmentation network and (ii) optimally balance the trade-off between false-negative and false-positive proposals.
Thus, we formulate a new practical training scheme, \textbf{Weakly Semi-Supervised Instance Segmentation (WSSIS) with point labels}.
In the WSSIS task, we utilize a small amount of fully labeled data and a massive amount of point labeled data for budget-efficient and high-performance instance segmentation.

Under the WSSIS setting, we filter out the proposals to keep only true-positive proposals using the point labels.
Then, given the true-positive proposals, we exploit the mask representation of the network learned by fully labeled data to produce high-quality pseudo instance masks.
For properly leveraging point labels, we consider the characteristics of the feature pyramid network (FPN)~\cite{(fpn)lin2017feature}, which consists of multi-level feature maps for multi-scale instance recognition.
Each pyramid level is trained to recognize instances of particular sizes, and extracting instance masks from unfit levels often causes inaccurate predictions, as shown in Figure \ref{fig:pyramid}.
However, since point labels do not have instance size information, we handle this using an effective strategy named Adaptive Pyramid-Level Selection.
We estimate which level is the best fit based on the reliability of the network ($i.e.,$ confidence score) and then adaptively produce an instance mask at the selected level.

Meanwhile, on an extremely limited amount of fully labeled data, the network often fails to sufficiently represent the instance mask in the mask branch, resulting in rough and noisy mask outputs.
In other words, the true-positive proposal does not always lead to a true-positive instance mask in this case.
To cope with this limitation, we propose a MaskRefineNet to refine the rough instance mask.
The MaskRefineNet takes three input sources, $i.e.,$ image, rough mask, and point;
the image provides visual information about the target instance, the rough mask is used as the prior knowledge to be refined, and the point information explicitly guides the target instance.
Using the richer instructive input sources, MaskRefineNet can be stably trained even with a limited amount of fully labeled data.

To demonstrate the effectiveness of our method, we conduct extensive experiments on the COCO~\cite{(coco)lin2014microsoft} and BDD100K~\cite{(bdd100k)yu2020bdd100k} datasets.
When training with half of the fully labeled images and the rest of the point labeled images on the COCO dataset ($i.e.,$ 50\% COCO), we achieve a competitive performance with the fully-supervised performance (38.8\% vs. 39.7\%).
In addition, when using a small amount of fully labeled data, $e.g.,$ 5\% of COCO data, the proposed method shows much superior performance than the state-of-the-art SSIS method~\cite{(NB)wang2022noisy} (33.7\% vs. 24.9\%).

\begin{figure*}[t]
    \centering
    \begin{subfigure}[b]{0.31\textwidth}
        \centering
        \includegraphics[width=\textwidth]{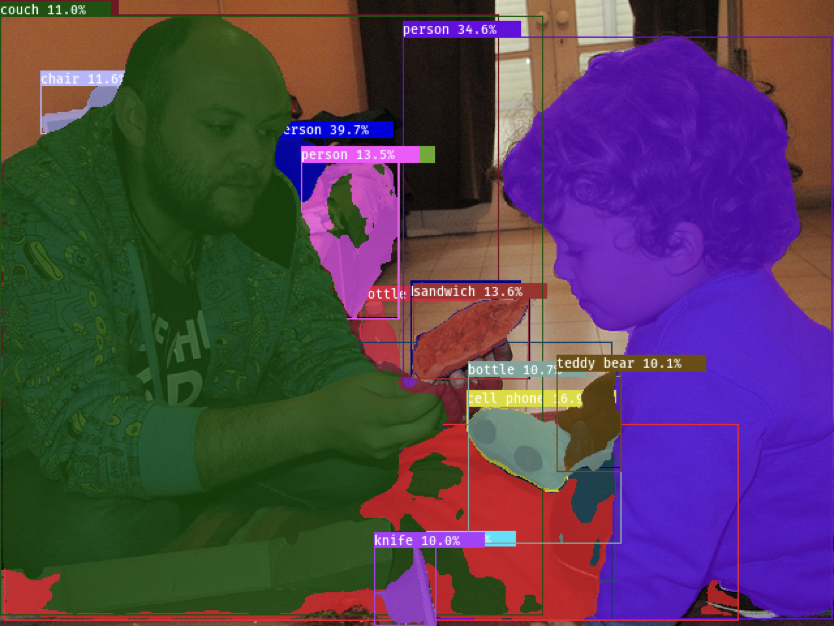}
        \caption{Confidence Threshold=0.1}
        \label{fig:quality_of_pseudo_label_semi_th01}
    \end{subfigure}
    \begin{subfigure}[b]{0.31\textwidth}
        \centering
        \includegraphics[width=\textwidth]{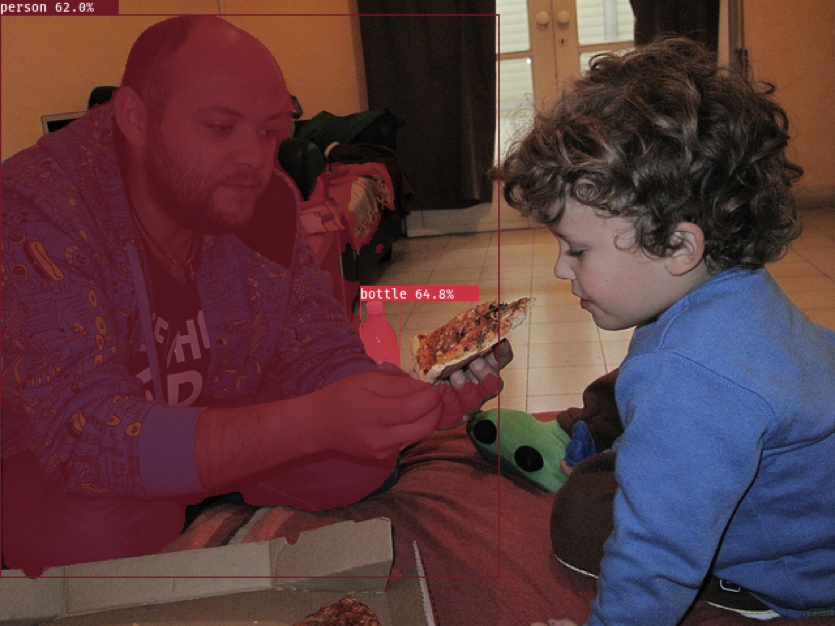}
        \caption{Confidence Threshold=0.5}
        \label{fig:quality_of_pseudo_label_semi_th05}
    \end{subfigure}
    \begin{subfigure}[b]{0.31\textwidth}
        \centering
        \includegraphics[width=\textwidth]{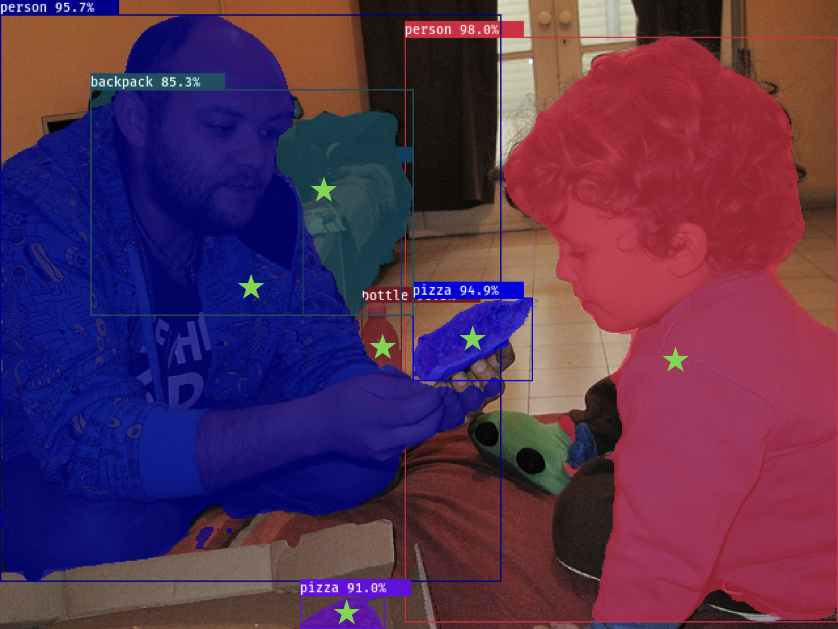}
        \caption{With Point Labels}
        \label{fig:quality_of_pseudo_label_weak_semi}
    \end{subfigure}
    \caption{
        \textbf{The qualitative results of pseudo instance masks}. (a) and (b): the quality of pseudo masks is largely affected by the confidence score of the proposal due to the trade-off between false-negative and false-positive instance proposals. (c): our point-driven method can filter the proposals to keep only true-positive proposals, resulting in clearer quality of pseudo instance masks.
    }
    \label{fig:quality_of_pseudo_label}
\end{figure*}

In summary, the contributions of our paper are
\begin{itemize}
    \item We show that point labels can be leveraged as effective weak supervisions for budget-efficient and high-performance instance segmentation. Further, based on this observation, we establish a new training protocol named Weakly Semi-Supervised Instance Segmentation (WSSIS) with point labels.
    \item To further boost the quality of the pseudo instance masks when the amount of fully labeled data is extremely limited, we propose the MaskRefineNet, which refines noisy parts of the rough instance masks.
    \item Extensive experimental results show that the proposed method can achieve competitive performance to those of the fully-supervised models while significantly outperforming the semi-supervised methods.
\end{itemize}
\section{Related Work}

\subsection{Instance Segmentation}
Mask R-CNN~\cite{(MRCNN)he2017mask} is the most widely used method for instance segmentation.
They represent an instance as a bounding box and produce the instance mask after pooling each box region.
These box-based approaches have many variants, such as \cite{(HTC)chen2019hybrid, (centermask)lee2020centermask,(msrcnn)huang2019mask,(detectors)qiao2021detectors} and have shown state-of-the-art results.
Meanwhile, there is a different type of approach, named point-based approaches~\cite{(solov2)wang2020solov2, (condinst)tian2020conditional,(centermask)wang2020centermask,(solo)wang2020solo}.
They represent an instance as a point and generate the instance mask using the point-encoded mask representation.
For example, SOLOv2~\cite{(solov2)wang2020solov2} extracts point-encoded kernel parameters and generates instance masks with a dynamic convolution scheme.
We note that the inference pipeline of these two approaches is the same as shown in Figure~\ref{fig:motivation}; they generate proposals in the form of either bounding boxes or points and then produce an instance mask for each proposal.
Here, the proposal is indispensable for producing the instance mask.

\subsection{Budget-Efficient Instance Segmentation}
Instance segmentation requires a huge amount of instance-level segmentation labels.
However, the annotation cost of segmentation labels is much higher than other labels. 
According to seminar works~\cite{(what_point)bearman2016s, (budget_aware)bellver2019budget}, the annotations time is measured on VOC dataset~\cite{(voc)everingham2010pascal} as follows: image-level (20.0 \textit{s/img}), point (23.3 \textit{s/img}), bounding box (38.1 \textit{s/img}), full mask (239.7 \textit{s/img}).
To reduce the annotation cost, weakly-supervised instance segmentation (WSIS) and semi-supervised instance segmentation (SSIS) have been actively studied.
The WSIS methods exploit the activation maps generated by self-attention of the network trained with only cost-efficient labels such as image-level~\cite{(irn)ahn2019weakly,(BESTIE)kim2022beyond,(prm)zhou2018weakly}, point~\cite{(BESTIE)kim2022beyond, cheng2022pointly,(wisenet)laradji2020proposal}, and bounding box~\cite{(BBAM)lee2021bbam, (boxinst)tian2021boxinst,(tightness)hsu2019weakly} labels.
Meanwhile, the SSIS methods~\cite{(NB)wang2022noisy,(shapeprop)zhou2020learning} use a small amount of fully labeled data and an abundant amount of unlabeled data.
Utilizing the knowledge of the segmentations learned with the fully labeled data, they generate pseudo instance masks for the unlabeled data.
Although they can reduce the annotation cost, their performance is still far behind those of the fully-supervised models.

\subsection{Weakly Semi-Supervised Object Detection}
There exist some previous attempts to tackle the weakly semi-supervised object detection problem using point labels (WSSOD)~\cite{(point_detr)chen2021points, (group_rcnn)zhang2022group}.
Namely, they use a few box-labeled data and a lot of point-labeled data.
Leveraging the point labels, they show improved detection performances compared to the semi-supervised setting.
Object detection and instance segmentation tasks share a similar goal: both are object-level recognition tasks.
However, we point out that the motivation for leveraging point labels is different. We focus on the fundamental drawback of the instance segmentation network to handle the trade-off between false-negative and false-positive proposals.
In contrast, PointDETR~\cite{(point_detr)chen2021points} leverages the point labels as input queries for single-level feature map inference of DETR~\cite{(detr)carion2020end} architecture, and Group R-CNN~\cite{(group_rcnn)zhang2022group} employs the point labels to filter and augment proposals with improved positive sample assignments.
In addition, we propose the MaskRefineNet for high-fidelity mask refinement to handle the distinct challenge of instance segmentation, which is a pixel-level recognition task.
\section{Proposed Method}

\subsection{Motivation}
Existing instance segmentation methods typically adopt a two-step inference process: (1) generate proposals where each instance is represented as bounding box~\cite{(MRCNN)he2017mask, (HTC)chen2019hybrid, (centermask)lee2020centermask,(msrcnn)huang2019mask} or point~\cite{(solov2)wang2020solov2, (condinst)tian2020conditional,(solo)wang2020solo,(centermask)wang2020centermask} in proposal branch, and (2) produce a mask for each instance in mask branch.
Figure \ref{fig:motivation} provides an intuition that the performance of the instance segmentation network critically depends on the correctness of proposals at the proposal branch.
Thus, improving the proposal branch may lead to a significant performance improvement in semi-supervised instance segmentation (SSIS).

To delve deeper into the problem,  we adjust a confidence threshold in the proposal branch to verify the influence of the proposal on the output instance mask as shown in Figure \ref{fig:quality_of_pseudo_label}.
At a low threshold of 0.1 with a larger number of proposals, we obtain more true-positive masks but much more false-positive masks as well (see Figure \ref{fig:quality_of_pseudo_label_semi_th01}).
The reason is that false-positive proposals ($e.g.,$ misclassified or erroneously localized proposals) often lead to noisy mask predictions.
Conversely, when we increase the threshold to 0.5, we lose several true-positive masks that were detected at lower thresholds (see Figure \ref{fig:quality_of_pseudo_label_semi_th05}).
In other words, although the mask branch could represent the instance mask, the absence of the thresholded proposal results in missing instance masks. However, finding an optimal threshold per instance is impractical, and balancing between true-positive and false-positive proposals still remains a challenging problem in SSIS.

\begin{figure}[t]
    \centering
    \begin{subfigure}[b]{\linewidth}
        \centering
        \includegraphics[width=\linewidth]{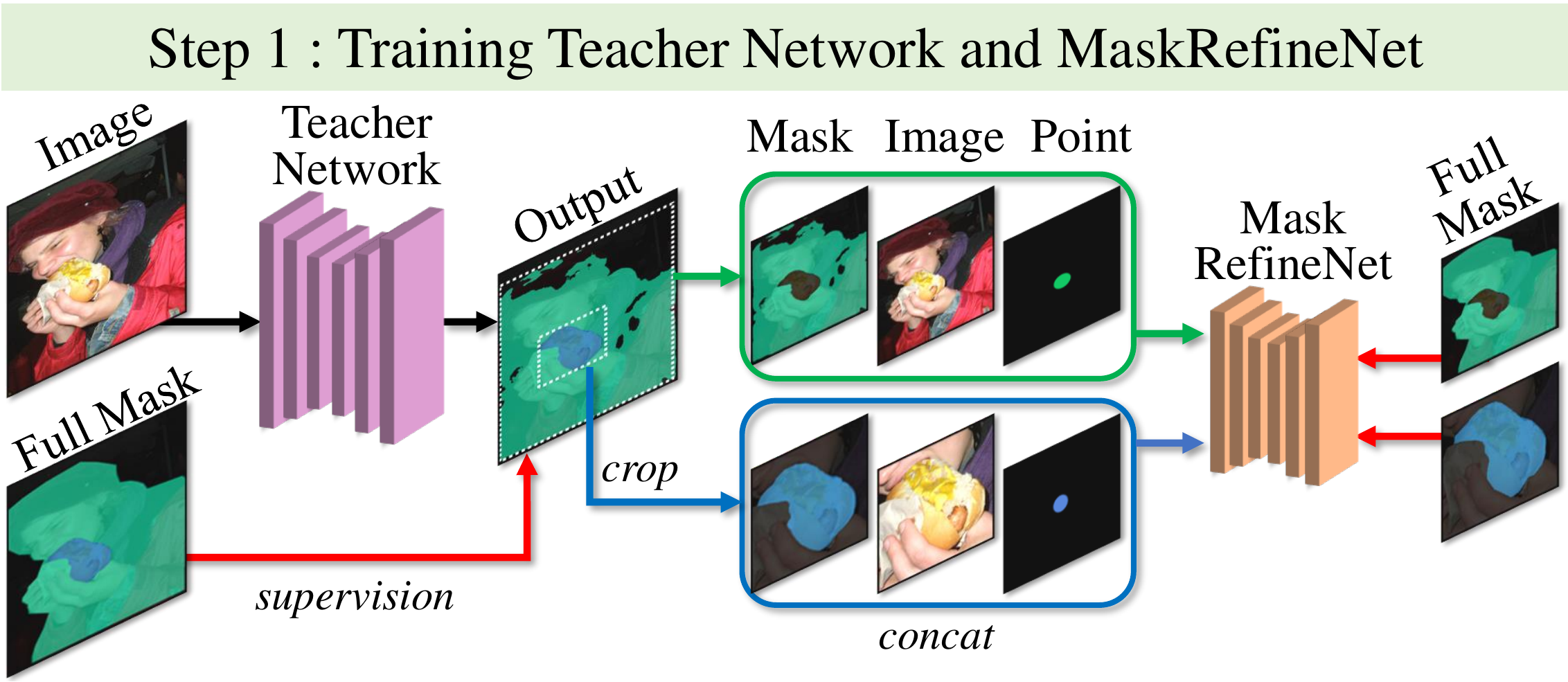} 
    \end{subfigure}
    \begin{subfigure}[b]{\linewidth}
        \centering
        \includegraphics[width=\linewidth]{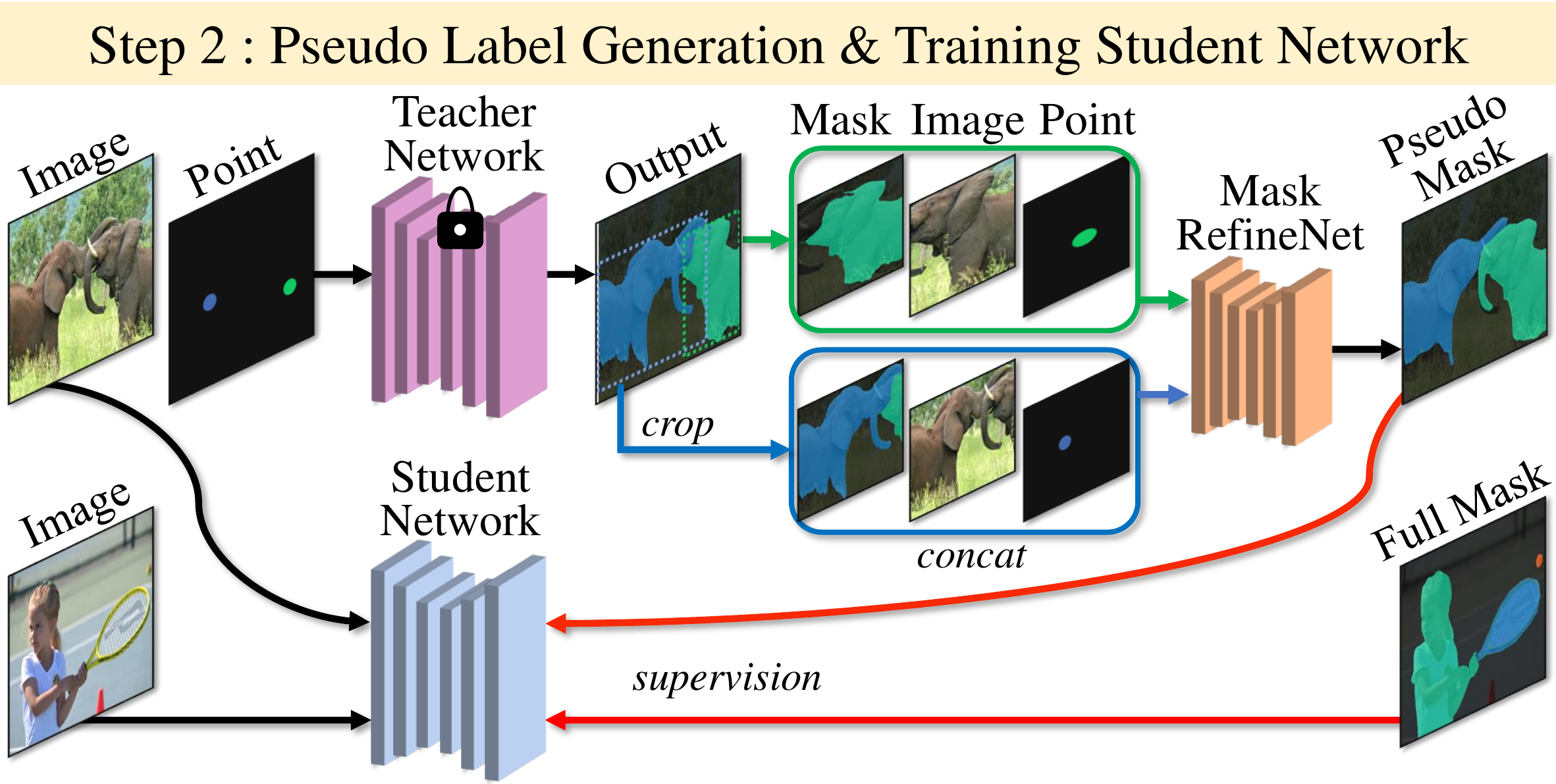} 
    \end{subfigure}
    \vspace{-2mm}
    \caption{
        \textbf{Overview of the proposed method}. Top: (step 1) training the teacher network and MaskRefineNet with fully labeled data. Bottom: (step 2) under the point label guidance, pseudo labels are generated through the teacher network and further refined using MaskRefineNet. Then, the student network is trained on both the pseudo-labeled data and the fully labeled data.
    }
    \label{fig:overview}
    \vspace{-2mm}
\end{figure}

\subsection{Weakly Semi-Supervised Instance Segmentation using Point Labels}

From the above observations, we can expect that obtaining correct instance proposals will yield accurate mask representations to improve an SSIS network.
To this end, we revisit the point label, which is a one-pixel categorical instance representation cue.
The annotation budget for point labels is known as costly-efficient by the literature~\cite{(budget_aware)bellver2019budget,(what_point)bearman2016s}.

\noindent\textbf{Task definition}. We propose a new training protocol named Weakly Semi-Supervised Instance Segmentation (WSSIS) using point labels and verify that the budget-friendly point labels can provide effective guidance. The training protocol employs point-labeled data with a small amount of fully labeled data, which yields reduced annotation costs.

\noindent\textbf{Training basline}. Figure \ref{fig:overview} shows our proposed baseline of a two-step learning procedure for WSSIS: (1) train a teacher network using only the full labels; (2) train a student network using both the full and pseudo labels generated by the teacher network along with the point labels.
Generating high-quality pseudo labels is crucial for WSSIS, so we employ point labels as guidance for filtering the proposals to remain only true-positive proposals.
Then, given the filtered proposals, we generate instance masks by exploiting the mask representation of the teacher network. Note that the proposed architecture is a baseline for the proposed task so that one can explore a more advanced training scheme.

\begin{figure}[t]
    \centering
    \begin{subfigure}[b]{\linewidth}
        \centering
        \includegraphics[width=0.96\linewidth]{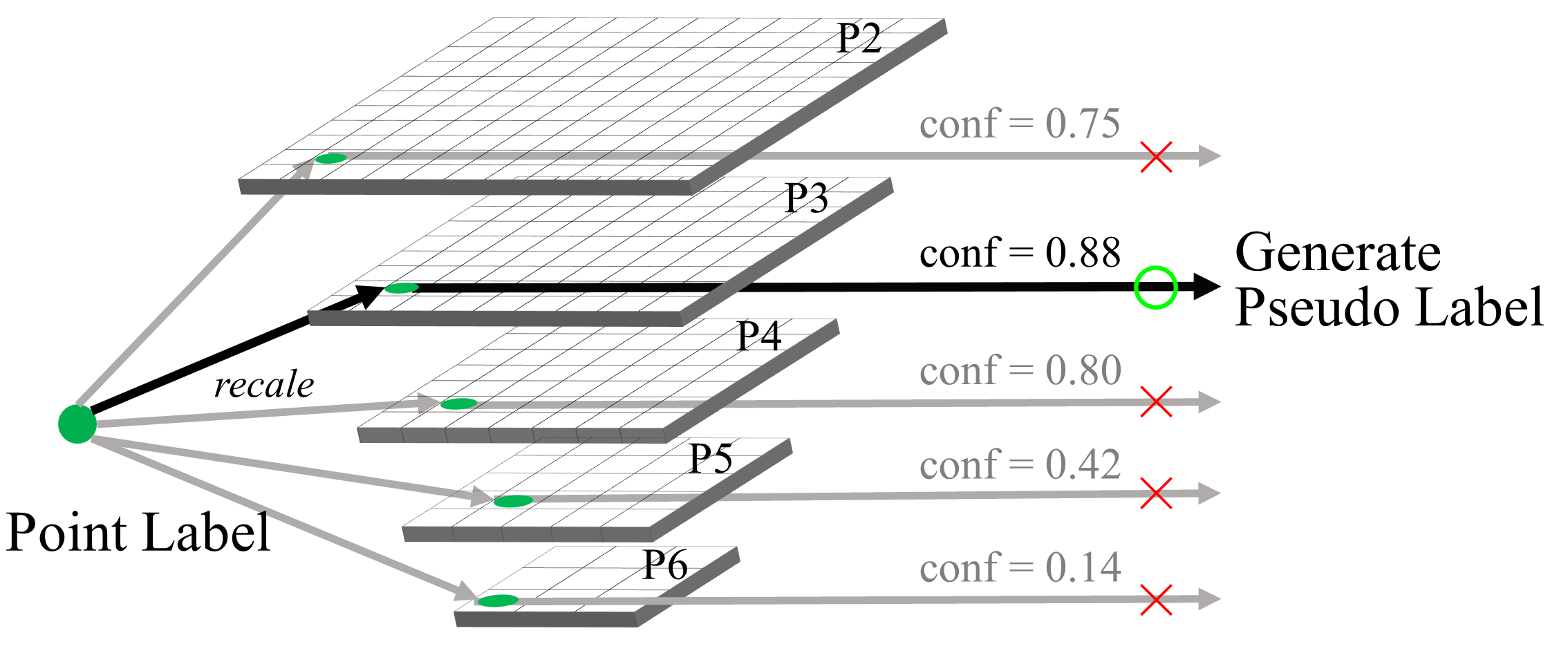} 
    \end{subfigure}
    \vspace{-2mm}
    \begin{subfigure}[b]{\linewidth}
        \centering
        \includegraphics[width=0.96\linewidth]{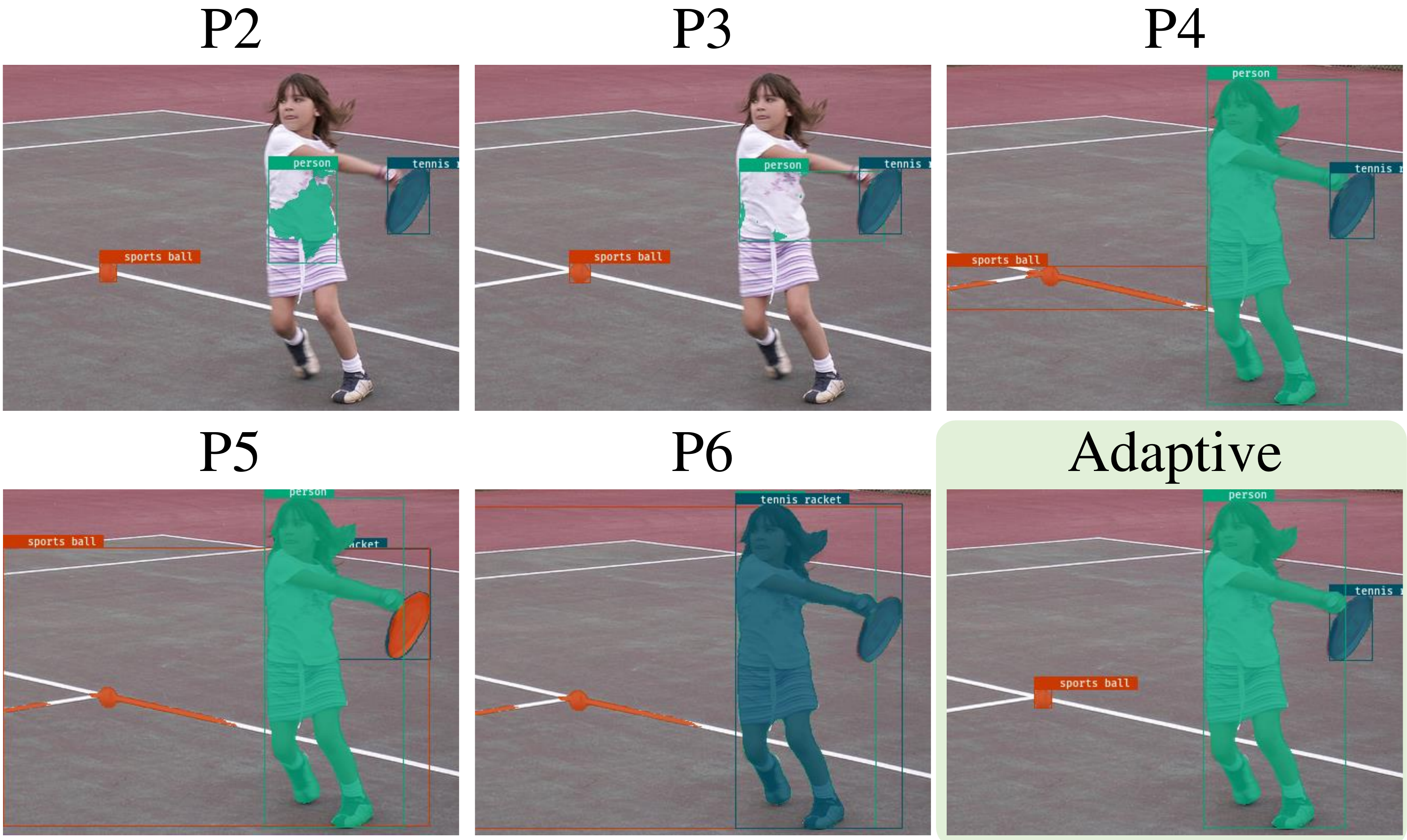} 
    \end{subfigure}
    \caption{
        \textbf{Adaptive strategy with FPN and qualitative results}. Top: illustration of Adaptive Pyramid-Level Selection. Bottom: the qualitative results from each pyramid level.
    }
    \label{fig:pyramid}
    \vspace{-2mm}
\end{figure}

\noindent\textbf{FPN head}.
Most existing instance segmentation approaches~\cite{(MRCNN)he2017mask,(solov2)wang2020solov2,(HTC)chen2019hybrid,(condinst)tian2020conditional} adopt Feature Pyramid Network (FPN)~\cite{(fpn)lin2017feature} architecture for multi-scale instance prediction.
Namely, SOLOv2~\cite{(solov2)wang2020solov2} employs a 5-level feature pyramid (P2$\sim$P6), and each pyramid level recognizes instances of particular sizes.
When combined with using point labels for sampling proposals, a careful approach to which level to extract proposals based on the size of the instance is demanding.
Otherwise, 
generated instance masks are often noisy as shown in Figure \ref{fig:pyramid} below.

\noindent \textbf{Strategy of using pyramid-level adaptively}. Since points do not contain instance size information, we estimate which pyramid level is proper for each point. To this end, we propose a strategy named Adaptive Pyramid-Level Selection, which adaptively selects a pyramid level that is expected to produce the most appropriate instance mask based on the reliability of the network.
Namely, we rescale the coordinate of point labels according to the resolution of each level and extract confidence scores for all levels.
Then, we generate an instance mask only from the pyramid level with the maximum confidence score, as illustrated in Figure \ref{fig:pyramid}.
Formally, there is $N$ proposal branches $\{\mathbf{F}^{p}_{i}\}_{i=1}^{N}$, and we follows the configuration of FPN~\cite{(fpn)lin2017feature} with $N{=}5$.
For each point label $(x, y, c)$, where $c$ denotes category id, we extract an instance proposal and confidence score $(\mathbf{P}_{i}, \mathbf{s}_{i})=\mathbf{F}^{p}_{i}(x, y, c)$.
Regarding the confidence score as the reliability of the prediction, we adaptively select a pyramid level $k$ with the maximum score, $k=\text{argmax}_{k{\in}\{1,2,\dots,N\}}\mathbf{s}_{k}$.
Finally, at the mask branch $\mathbf{F}^{m}$, we generate a pseudo instance mask $M={\sigma}(\mathbf{F}^{m}(\mathbf{P}_{k}))$, where $\sigma$ is sigmoid function.

\begin{figure}[t]
    \centering
    \includegraphics[width=0.9\linewidth]{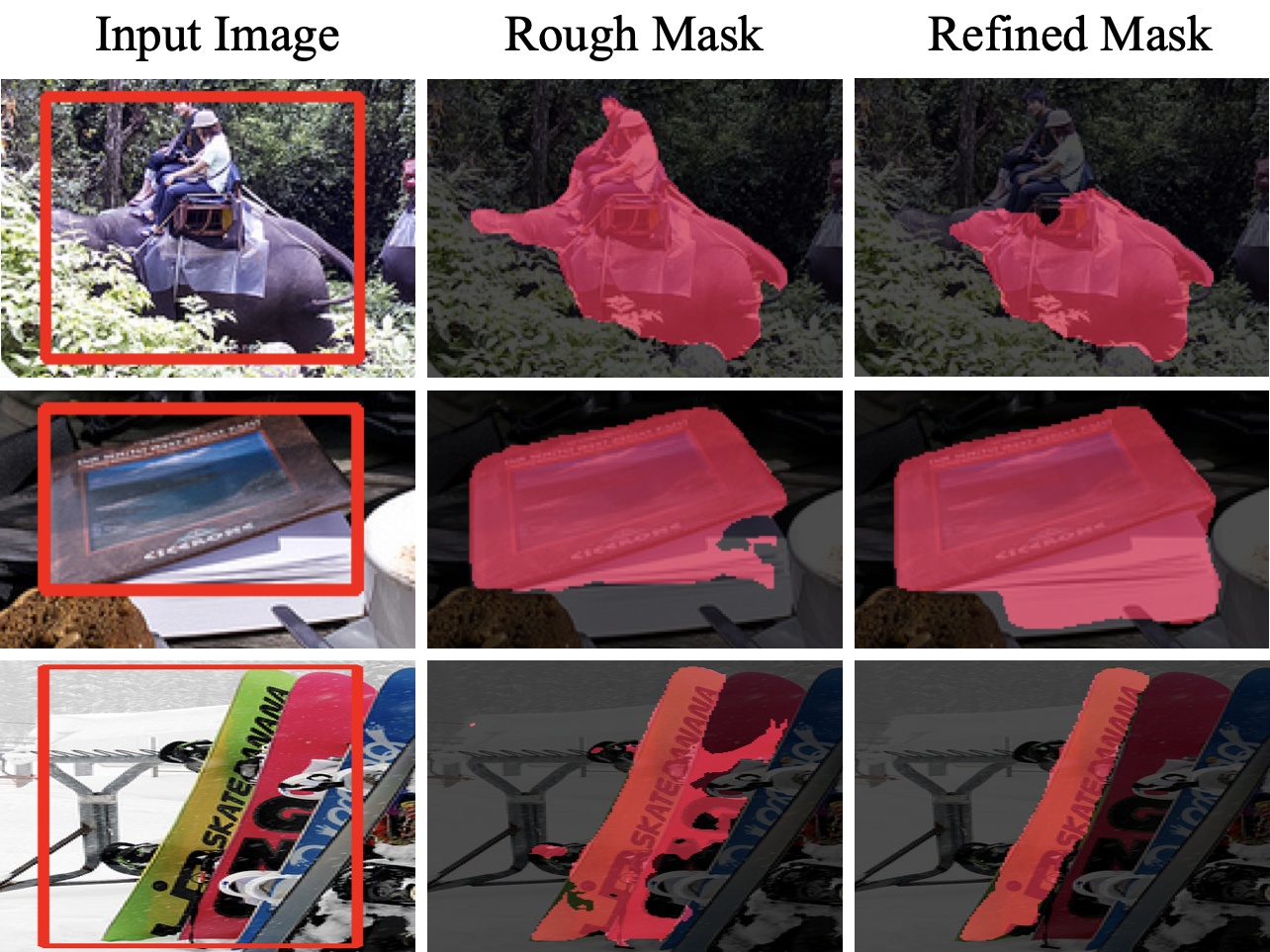}
    \caption{
        \textbf{Effect of MaskRefineNet}. The qualitative results under 10\% of COCO fully labeled data condition. When the teacher network fails to disentangle objects in a rough mask, MaskRefineNet can separate each representation owing to the given point label (1st row). Our MaskRefineNet further enriches the resultant mask representation (2nd row) and removes noisy parts (3rd row).
    }
    \label{fig:quality_of_maskrefinenet}
    \vspace{-2mm}
\end{figure}

\subsection{Mask Refinement Network}

With a sufficient amount of fully labeled data ($e.g.,$ using 50\% images), the teacher network can afford to generate reasonable pseudo-instance masks given true-positive proposals.
However, when the amount of fully labeled data is extremely small ($e.g.,$ using only 1\% images), the mask representation by the network would produce rough instance masks; it means the true-positive proposal could not ensure that the instance mask is a true-positive.

To handle such a challenging case, we propose a simple yet effective post-hoc mask refinement method named MaskRefineNet.
Figure \ref{fig:overview} shows that MaskRefineNet can refine the rough mask output from the teacher network based on three input sources, including input image, rough mask, and point information. 
Specifically, we loosely crop each instance region in the input image, rough mask, and point information, and resize them to 256$\times$256, then concatenate them together into an input tensor.
For the point information, we transform the point label to the form of a heatmap where each point is encoded into a 2D gaussian kernel with a sigma of 6.
The effectiveness of the MaskRefineNet can be attributed to two reasons;
(1) it leverages the prior knowledge of the teacher network; since MaskRefineNet takes the rough mask predictions from the teacher network as the input, it learns how to calibrate common errors of predictions from the teacher network;
(2) it takes guidance from the input point that is likely to provide an accurate target instance cue for recognizing overlapping instances and falsely predicted pixels.
Consequently, MaskRefineNet refines the missing \& noisy parts and disentangles the crowded target instances in the rough mask as shown in Figure \ref{fig:quality_of_maskrefinenet} with the help of the point guidance.
\section{Experiments}

\subsection{Datasets}
We evaluate our method on the COCO 2017 dataset~\cite{(coco)lin2014microsoft} that contains 118,287 training samples and 5,000 validation samples for 80 common object categories.
To validate our method under the WSSIS regime, we randomly sample subsets containing 1\%, 2\%, 5\%, 10\%, 20\%$\sim$50\% of the COCO training dataset.
COCO 10\% means using 10\% of the fully labeled data and the rest of 90\% of the point labeled data.
We use a centroid point of an instance mask label as a point label.
In addition, we conduct experiments on BDD100K dataset~\cite{(bdd100k)yu2020bdd100k}, which is a large-scale driving scene dataset with diverse scene types and 8 classes.
The BDD100K dataset contains 7k mask-labeled images and 67k box-labeled images, and we use the center of the box as the point label for this dataset.

\subsection{Implementation Details}
We adopt SOLOv2~\cite{(solov2)wang2020solov2} as the baseline instance segmentation network since it is a point-based and box-free straightforward method.
For both teacher and student networks, we use the same ResNet-101~\cite{(resnet)he2016deep} backbone network and follow the default training recipe and network setting as in \cite{(solov2)wang2020solov2}.
For the MaskRefineNet, we adopt the ResNet-101 FPN~\cite{(fpn)lin2017feature} architecture and produce the output only from the highest resolution pyramid level, P2.
We set the batch size of 16, the learning rate of 1e-4 with cosine decay scheduling, dice loss~\cite{(dice_loss)milletari2016v}, and input size of 256$\times$256 for training the MaskRefineNet.
After training the teacher network, the MaskRefineNet is trained by taking the rough mask outputs from the teacher network.
We implement the proposed method using Pytorch~\cite{(pytorch)paszke2019pytorch} and train on 8 V100 GPUs.

Following the labeling budget calculation in \cite{(what_point)bearman2016s, (budget_aware)bellver2019budget}, we estimate the labeling budget for the COCO trainset as follows: Full mask ($645.9s/img$), Bounding box ($127.5s/img$), Point ($87.9s/img$), Image-level ($80s/img$).
Detailed calculation method is described in our supplementary material.

\begin{figure}[t]
    \centering
    \includegraphics[width=\linewidth]{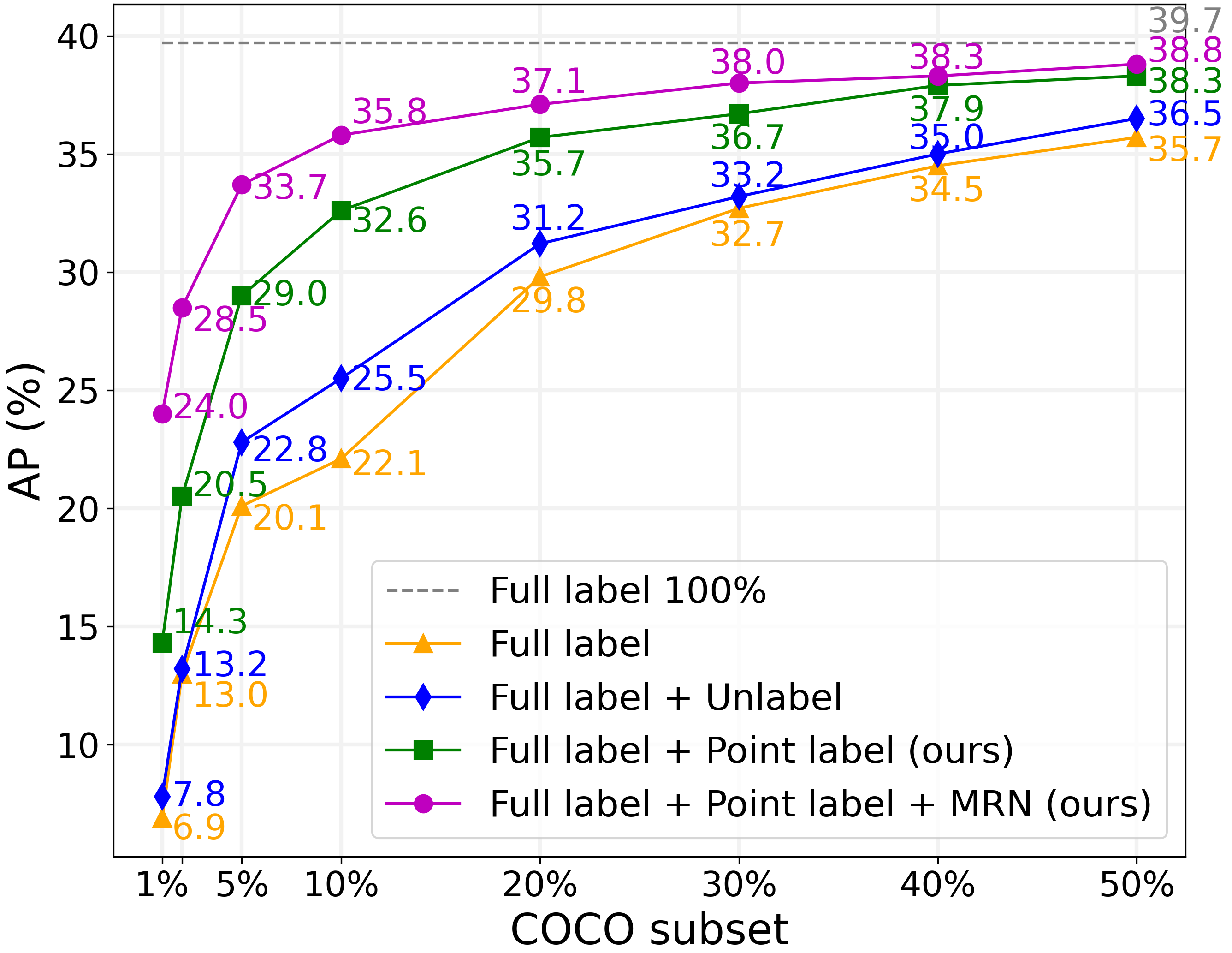}
    \caption{
        \textbf{Performance trend comparison under supervisions}. We visualize the AP scores when varying numbers of fully labeled data in the COCO test-dev. MRN means applying our point-guided MaskRefineNet.
    }
    \label{fig:result_coco}
\end{figure}

\subsection{Experimental Results}

\begin{figure*}[t]
    \centering
    \includegraphics[width=1\linewidth]{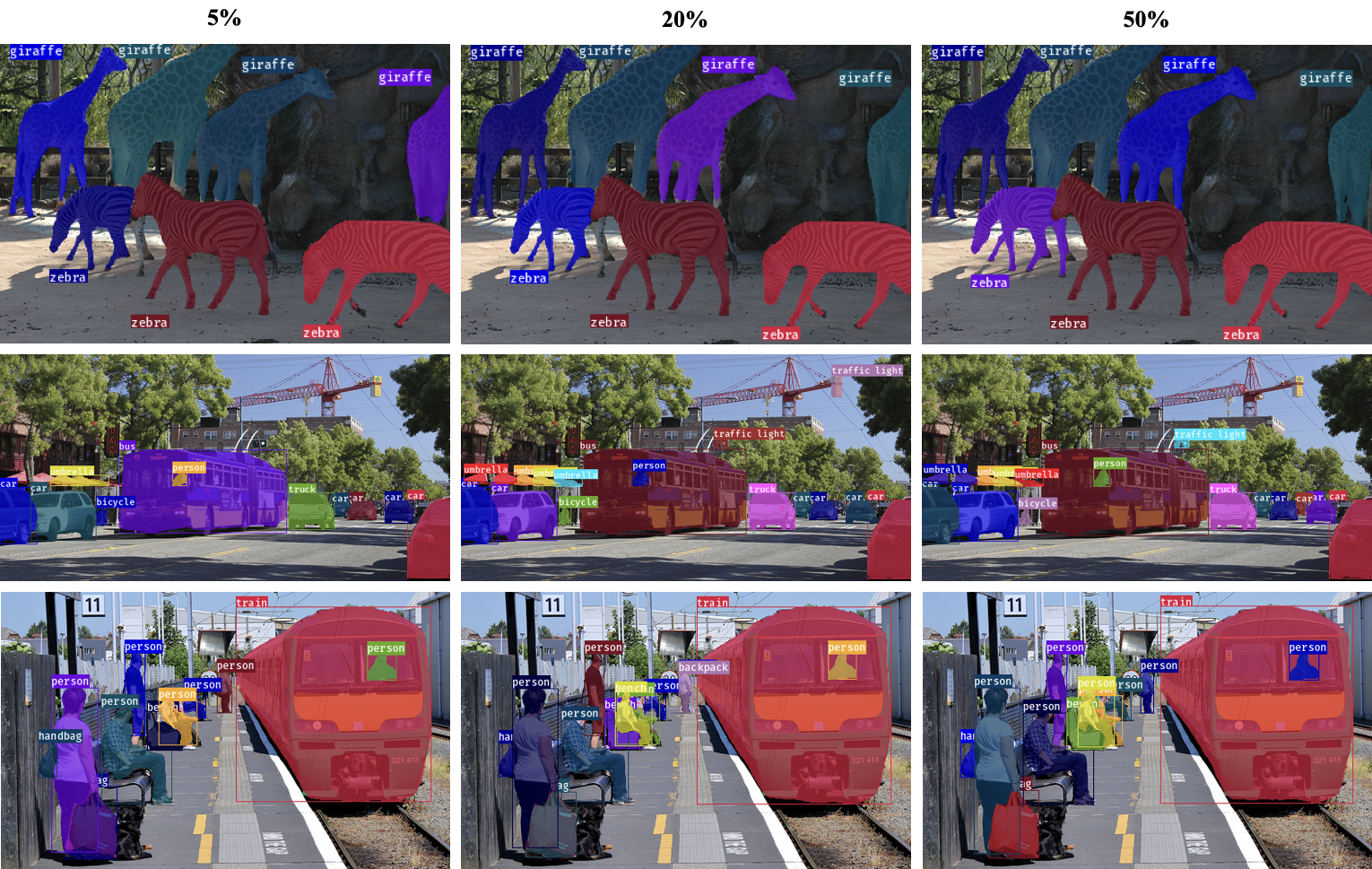}
    \caption{
        \textbf{Qualitative results according to the various subsets in COCO data}. We observe that training with 5\% of full labeled data appropriately localizes all the object instance masks owing to our proposed point guidance method along with MaskRefineNet.
    }
    \label{fig:abs_coco_percent}
    \vspace{-1mm}
\end{figure*}

We compare the proposed method against two baselines with the same network architecture and optimization strategy.
The first is training with only fully labeled data, and the second is training with fully labeled and unlabeled data, which is a semi-supervised setting.
For the second baseline, we generate pseudo instance masks for the unlabeled data from the teacher network without any weak labels.
As shown in Figure \ref{fig:result_coco}, our method achieves remarkable performances on all COCO subsets.
Especially, the performance gap between ours and the baselines is notably larger when we use smaller subsets with fully labeled images, $e.g.,$ COCO 1\% or 5\%.
Compared to the fully-supervised setting (COCO 100\%), our method with COCO 50\% shows a highly competitive result (38.8\% vs. 39.7\%).
Moreover, the qualitative results in Figure \ref{fig:abs_coco_percent} show that ours with COCO 5\% can properly segment instances of various sizes.
These results demonstrate that the cost-efficient point labels can be leveraged as an effective source for instance segmentation.

We also compare against other methods that use various weak labels based on the labeling budget in Table \ref{tab:comparison_budget}.
According to the type and the number of labels, we calculate the labeling budget following the aforementioned cost.
All methods use the same amount of total training images, so the labeling time of unlabeled data is treated as zero.
Compared to the state-of-the-art semi-supervised method, NB~\cite{(NB)wang2022noisy}, ours show superior performances when using the same amount of fully labeled data, especially when using 5\% of fully labeled data (33.7\% vs. 25.6\%).
We also achieve higher performance with a lower labeling budget (35.8\% with a budget of 196.7 vs. 35.5\% with a budget of 265.2).
In addition, compared to the state-of-the-art box-supervised method, BoxInst~\cite{(boxinst)tian2021boxinst}, our efficiency is better (33.7\% with a budget of 158.5 vs. 33.2\% with a budget of 174.5).
This result demonstrates the effectiveness of budget-friendly point labels with the proposed method.
Furthermore, we emphasize the potential of our method for performance improvement when using more labeling budget.

\begin{table}[t]
  \centering
  \small
  \begin{adjustbox}{max width=\linewidth}
  \begin{tabular}{c|c|c|c}
    \toprule
    Method & Label Types & Budget (days) $\downarrow$ & $AP$ (\%) $\uparrow$ \\
    \midrule %
    \multicolumn{4}{c}{\footnotesize \textit{\textbf{Weakly-supervised Models $\downarrow$}}} \\
    \midrule
    BESTIE~\cite{(BESTIE)kim2022beyond} & $\mathcal{I}$ 100\% & 109.5 & 14.3  \\
    BESTIE~\cite{(BESTIE)kim2022beyond} & $\mathcal{P}$ 100\% & 120.3 & 17.7  \\
    BBAM~\cite{(BBAM)lee2021bbam} & $\mathcal{B}$ 100\% & 174.5 & 26.0  \\
    BoxInst~\cite{(boxinst)tian2021boxinst} & $\mathcal{B}$ 100\% & 174.5 & 33.2  \\
    \midrule
    \multicolumn{4}{c}{\footnotesize 
 \textit{\textbf{Semi-supervised Models $\downarrow$}}} \\
    \midrule
    NB~\cite{(NB)wang2022noisy} & $\mathcal{F}$ 5\% + $\mathcal{U}$ 95\% & 44.2 & 25.6  \\
    NB~\cite{(NB)wang2022noisy} & $\mathcal{F}$ 10\% + $\mathcal{U}$ 90\% & 88.4 & 30.3  \\
    NB~\cite{(NB)wang2022noisy} & $\mathcal{F}$ 30\% + $\mathcal{U}$ 70\% & 265.2 & 35.5  \\
    NB~\cite{(NB)wang2022noisy} & $\mathcal{F}$ 50\% + $\mathcal{U}$ 50\% & 442.1 & 36.8  \\
    \midrule
    \multicolumn{4}{c}{\footnotesize \textit{\textbf{Weakly Semi-supervised Models $\downarrow$}}} \\
    \midrule
    Ours & $\mathcal{F}$ 5\% + $\mathcal{P}$ 95\%  & 158.5 & 33.7  \\
    Ours & $\mathcal{F}$ 10\% + $\mathcal{P}$ 90\% & 196.7 & 35.8  \\
    Ours & $\mathcal{F}$ 30\% + $\mathcal{P}$ 70\% & 349.5 & 38.0  \\
    Ours & $\mathcal{F}$ 50\% + $\mathcal{P}$ 50\% & 502.3 & 38.8  \\
    \midrule
    \multicolumn{4}{c}{\footnotesize \textit{\textbf{Fully Supervised Models $\downarrow$}}} \\
    \midrule
    MRCNN~\cite{(MRCNN)he2017mask} & $\mathcal{F}$ 100\% & 884.2 & 38.8  \\
    SOLOv2~\cite{(solov2)wang2020solov2} & $\mathcal{F}$ 100\% & 884.2 & 39.6  \\
    \bottomrule
  \end{tabular}
  \end{adjustbox}
  \caption{
    \textbf{Performance trade-off of annotation budgets and AP}. We compare the methods on the COCO test-dev under various supervisions; $\mathcal{U}$ (unlabeled data), $\mathcal{I}$ (image-level label), $\mathcal{P}$ (point label), $\mathcal{B}$ (box label), $\mathcal{F}$ (full label). All methods use the same backbone network of R-101~\cite{(resnet)he2016deep}
  }
  \label{tab:comparison_budget}
\end{table}
\begin{table*}[t]
    \begin{minipage}[b]{0.48\linewidth}
        \centering
        \small
        \begin{adjustbox}{max width=1\linewidth}
        \begin{tabular}{c|c|ccc}
        \toprule
        FPN & Proposal & $AP$↑ & $AP_{50}$↑ & $AR_{100}$↑ \\
        \midrule
        P2 & P2 & 23.4 & 48.3 & 37.6 \\
        P2$\sim$P6 & P2$\sim$P6 & 10.3 & 20.8 & 44.7 \\
        P2$\sim$P6 & $ \footnotesize \text{argmax}_{k{\in}\{2,3,\dots,6\}}\mathbf{s}_{k}$ & 28.6 & 56.7 & 42.6 \\
        \midrule
        P2$\sim$P6 & \footnotesize \text{w/ ground-truth size} & 30.9 & 62.0  & 44.9 \\
        \bottomrule
        \end{tabular}
        \end{adjustbox}
        \caption{
            \textbf{Impact of choosing features adaptively}. P$n$ denotes $n$-th feature pyramid. $\mathbf{s}_{i}$ is the confidence score of the proposal in $i$-th pyramid level.
        }
        \label{tab:ablation_pyramid}
    \end{minipage}
    \hfill 
    \begin{minipage}[b]{0.27\linewidth}
        \centering
        \small
        \tabcolsep=0.1cm
        \begin{adjustbox}{max width=\linewidth}
        \begin{tabular}{cc|cc}
        \toprule
        Rough mask & Point & $AP$↑ & $AP_{50}$↑ \\
        \midrule
        \multicolumn{2}{c|}{\footnotesize \text{w/o MaskRefineNet}} & 28.6 & 56.7 \\
        \midrule
                   & & 14.8 & 30.2 \\
        \checkmark & & 29.7 & 54.4 \\
         & \checkmark & 30.9 & 52.9 \\
        \checkmark & \checkmark & 39.1 & 65.3 \\
        \bottomrule
        \end{tabular}
        \end{adjustbox}
        \caption{
            \textbf{Impat of the input sources for MaskRefineNet}.
        }
        \label{tab:ablation_maskrefinenet}
    \end{minipage}
    \ 
    \begin{minipage}[b]{0.2\linewidth}
        \centering
        \small
        \tabcolsep=0.15cm
        \begin{adjustbox}{max width=\linewidth}
        \begin{tabular}{c|cc}
        \toprule
        Point & $AP$↑ & $AP_{50}$↑  \\
        \midrule
        Center   & 28.6 & 56.7  \\
        Random   & 28.8 & 57.0 \\
        \bottomrule
        \end{tabular}
        \end{adjustbox}
        \caption{
            \textbf{Robustness to point sources}. We compare APs trained with different locations -- center and random points.
        }
        \label{tab:robustness_point}
        \vspace{1.8mm}    
    \end{minipage}
\end{table*}

\begin{table}[t]
  \centering
  \small
  \begin{tabular}{c|ccc|cc}
    \toprule
    \multirow{2}{*}{Label Types} & \multicolumn{3}{c|}{COCO \textit{train5K}} & \multicolumn{2}{c}{COCO \textit{val}} \\
    \cline{2-6}
     & $AP$↑ & $AP_{50}$↑ & $AR_{100}$↑ & $AP$↑ & $AP_{50}$↑  \\
    \midrule
    $\mathcal{U}$ ($\tau$=0.1)    & 6.0 & 11.3  & 33.8 & 20.8 & 33.5 \\
    $\mathcal{U}$ ($\tau$=0.3)    & 13.1 & 22.9 & 23.4 & 25.9 & 41.5 \\
    $\mathcal{U}$ ($\tau$=0.5)    & 12.2 & 19.3 & 15.6 & 24.3 & 38.2 \\
    $\mathcal{I}$ ($\tau$=0.3)    & 19.5 & 33.1 & 24.1 & 29.5 & 48.9 \\
    \midrule
    $\mathcal{P}$ & 28.6 & 56.7 & 42.6 &  32.2 & 52.3 \\
    $\mathcal{P}^{\dagger}$ & 39.1 & 65.3 & 52.0 & 35.5 & 56.0 \\
    \bottomrule
  \end{tabular}
  \caption{
    \textbf{Impact of using point labels}. We have the notations: $\mathcal{U}$ (unlabeled data), $\mathcal{I}$ (image-level label), $\mathcal{P}$ (point label), and $\mathcal{F}$ (full label).
    We use COCO \textit{train5K} to measure the quality of pseudo labels and COCO \textit{val} to evaluate the baseline network trained with the pseudo labels.
    $\tau$ is a confidence threshold in the proposal branch. $\dagger$ means applying our point-guided MaskRefineNet.
  }
  \label{tab:ablation_threshold}
\end{table}

Also, we conduct experiments on BDD100K dataset.
As we increase the amount of point labeled data with a fixed amount of 7k fully labeled data, the performance is gradually improved, as in Table \ref{tab:result_BDD100K}.
Especially, when leveraging all available point labels (67k), ours can achieve significant performance improvements compared to using only 7k fully labeled data (22.1\%{$\rightarrow$}27.9\%).

\subsection{Ablation Study}
We conduct an ablation study of our method on the COCO 10\% setting.
Unless otherwise specified, we measure the quality of pseudo labels generated by the teacher network using randomly sampled 5,000 images in the rest 90\% of COCO data, we name it COCO \textit{train5K}.

\noindent \textbf{Effect of Point Labels.} 
In Table \ref{tab:ablation_threshold}, we verify the effectiveness of each weak label candidate ($i.e.,$ unlabeled, image-level, and point label) in instance segmentation.
For this analysis, we measure the quality of pseudo labels and the performance of the student network on the COCO 2017 validation set.
When the unlabeled data is leveraged as a weak label, we should carefully tune the confidence threshold to balance between false-negative and false-positive proposals; the average recall ($AR_{100}$) and precision ($AP$) largely vary according to the confidence threshold.
It implies that human effort for tuning the threshold is required for target datasets, and this global threshold may not be optimal for every instance.
Leveraging the image-level label as a weak label can eliminate the misclassified proposals, boosting the performance from 25.9\% to 29.5\%.
However, the performance gap with the fully-supervised setting is still significant (29.5\% vs. 39.0\%).
When we leverage the point label as a weak label, we filter out the proposals to keep only true-positive proposals, deprecating the requirement of the confidence threshold.
It makes a more straightforward and effective pipeline, resulting in 32.2\%.
Compared to the annotation cost of the image-level label (80 $s/img$), the point label is still budget-friendly (87.9 $s/img$) and gives a noticeable performance improvement (32.2\% vs. 29.5\%).
Moreover, our point-guided MaskRefineNet further reduces the performance gap with the fully-supervised setting (35.5\% vs. 39.0\%).
This result demonstrates that our method can effectively leverage the point label for cost-efficient and high-performance instance segmentation.

Furthermore, we test the robustness of our method to the position of the point label.
We originally used the centroid point of each instance as our point label. 
For the analysis, we randomly choose one pixel in an instance mask as a point label five times and measure the average quality of the pseudo labels.
As shown in Table \ref{tab:robustness_point}, the performance gap between the center point and the random point is marginal.
The reason is that all pixels included in the instance region within the proposal branch are trained to generate instance proposals, as in \cite{(solov2)wang2020solov2}.
This result demonstrates the robustness of our method to the position of the point labels, which gives us more opportunity to reduce the annotation effort.

\begin{table}[t]
  \centering
  \small
  \begin{tabular}{c|ccc}
    \toprule
    Label Types & $AP$↑ & $AP_{50}$↑ & $AP_{75}$↑ \\
    \midrule
    $\mathcal{F}$ 7k & 22.1 & 40.2 & 21.2 \\
    $\mathcal{F}$ 7k + $\mathcal{P}$ 20k & 26.7 & 44.4 & 27.8 \\
    $\mathcal{F}$ 7k + $\mathcal{P}$ 40k & 27.3 & 44.5 & 28.9 \\
    $\mathcal{F}$ 7k + $\mathcal{P}$ 67k & 27.9 & 44.8 & 29.2 \\
    \bottomrule
  \end{tabular}
  \caption{
   \textbf{ Quantitative results on BDD100K validation set}. We report the AP scores with different training regimes concerning the number of point labels.
  }
  \label{tab:result_BDD100K}
\end{table}

\noindent \textbf{Effect of Adaptive Pyramid-Level Selection.} 
We quantitatively analyze the behavior of the FPN in Table \ref{tab:ablation_pyramid}.
When we produce pseudo instance masks from a single layer feature map, $i.e.,$ without FPN, we achieve an unsatisfactory pseudo label quality of 23.4\% as shown in the first row in Table \ref{tab:ablation_pyramid}.
When we generate pseudo masks from all pyramid feature maps (P2$\sim$P6), we achieve an inferior quality of 10.3\% because the outputs from unfit pyramid levels are pretty noisy, as shown in Figure \ref{fig:pyramid}.
Using our Adaptive Pyramid-Level Selection strategy, we choose one appropriate pyramid level based on the reliability of the network, achieving the improved quality of 28.6\%.
The result demonstrates that the proposed strategy is highly effective in leveraging the behavior of the FPN structure for generating high-quality pseudo labels.
Also, the result of 30.9\% when using ground-truth instance size information leaves us room for improvement of our method.

\noindent \textbf{Effect of MaskRefineNet.}
In Figure \ref{fig:result_coco}, we conduct experiments on various subsets without the MaskRefineNet.
When the teacher network has enough mask representation ability as in the COCO 50\% setting, the improvement of the MaskRefineNet is marginal (38.3\% vs. 38.8\%).
However, the MaskRefineNet yields a considerable performance improvement, especially in the limited number of fully labeled data settings, $e.g.,$ COCO 1\% (14.3\%{$\rightarrow$}24.0\%) and COCO 5\% (29.0\%{$\rightarrow$}33.7\%) settings. 
The result demonstrates that MaskRefineNet is a remarkably effective method to improve the quality of pseudo labels in the limited quantity of fully labeled data conditions.

In addition, we analyze the effect of input sources of the MaskRefineNet in Table~\ref{tab:ablation_maskrefinenet}.
Before applying the MaskRefineNet, the quality of pseudo labels is measured as 28.6\%.
When the MaskrefineNet only takes an image as input, the accuracy of the pseudo labels drastically reduces to 14.8\% since the network fails to converge due to the absence of prior knowledge.
When taking the rough mask as an input source, the quality of the pseudo labels improves from 14.8\% to 29.7\% because the MaskRefineNet takes the knowledge of the teacher network for fast and stable convergence.
However, the improvement is still minor compared to the model without MaskRefineNet.
When additionally taking the point information as an input source, the quality of pseudo labels dramatically improves to 39.1\%.
The reason is that the point information is used as a guidance seed for the target instance, helping a more accurate segment of occluded instances and refining the missing predictions in the rough mask, as shown in Figure \ref{fig:quality_of_maskrefinenet}.
\section{Conclusion and Limitation}

In this paper, we proposed a novel and practical weakly semi-supervised instance segmentation scheme leveraging point labels as weak supervision for cost-efficient and high-performance instance segmentation.
We motivated that the main performance bottleneck of modern instance segmentation frameworks arises from the instance proposal extraction.
To this end, we proposed a method that can effectively exploit the budget-friendly point labels as weak supervision to resolve the bottleneck.
Moreover, we presented the MaskRefineNet to deal with hard learning scenarios where the amount of fully labeled data is extremely limited.
Owing to the effectiveness of the proposed method, we can generate high-quality pseudo instance masks, achieving promising instance segmentation results.
Despite our remarkable results driven by cost-efficient point labels, we have a limit to straightforwardly exploiting the tremendous amount of unlabeled image pools, such as web-crawling images without any annotations.
Our future direction may involve incorporating unlabeled images in our framework to extend its application to semi-supervised learning scenarios.

\paragraph{Acknowledgment.} We thank Youngjoon Yoo and ImageVision team members for valuable discussions and feedback. This work was supported by the Engineering Research Center Program through the National Research Foundation of Korea (NRF) funded by the Korean Government MSIT (NRF-2018R1A5A1059921) and Institute of Information \& communications Technology Planning \& Evaluation (IITP) grant funded by the Korea government (MSIT) (No.2019-0-00075, Artificial Intelligence Graduate School Program (KAIST)).

\clearpage

\section*{Appendix: Additional Experimental Details}

\paragraph{Labeling Budget Calculation.}
Seminar works~\cite{(what_point)bearman2016s, (budget_aware)bellver2019budget} offered the annotation time of various labeling sources ($e.g.,$ full mask, bounding box, point, image-level labels) on Pascal VOC dataset~\cite{(voc)everingham2010pascal}.
Since the COCO dataset~\cite{(coco)lin2014microsoft} we used has more categories and instances per image than the VOC dataset, we estimate the labeling budget for the COCO dataset following their budget calculation method.
The COCO 2017 trainset has a total of 80 categories and contains 118,287 images and 860,001 instances.
Also, it has an average of 7.2 instances and 2.9 categories per image.
By considering this statistic of COCO dataset, we calculate the labeling budget as follows: 
\begin{itemize}
    \item \textbf{Full mask}: $77.1 classes/img \times 1s/class+7.2inst/img \times 79s/mask=\textbf{645.9s/img}$.
    \item \textbf{Bounding box}: $77.1 classes/img \times 1s/class+7.2inst/img \times 7s/bbox=\textbf{127.5s/img}$.
    \item \textbf{Point}: $77.1 classes/img \times 1s/class+2.9 classes/img \times 2.4s/point+(7.2inst/img-2.9 classes/img) \times 0.9s/point=\textbf{87.9s/img}$.
    \item \textbf{Image-level}: $80 classes/img \times 1s/class=\textbf{80s/img}$.
\end{itemize}

\paragraph{Input of MaskRefineNet.}
In this section, we further provide the details about the input sources for MaskRefineNet.
After training the teacher network using the fully labeled data, we generate instance mask outputs for the point-guided filtered proposals ($i.e.,$ true-positive proposals) using the trained teacher network.
We treat the mask outputs as rough masks to be used as the input source of the MaskRefineNet.
For each rough mask, we loosely crop each instance region in the input image, rough mask, and point heatmap.
Specifically, after obtaining the bounding box from the rough mask using the min-max operations, we re-scale the size of the box to double, and then we use this box region as the cropping region.
In addition, for the point heatmap, we encode each point to a 2-dimensional gaussian kernel with a sigma of 6, as done in ~\cite{(centermask)wang2020centermask,(centernet)zhou2019objects}.
We concatenate the three input sources ($i.e.,$ cropped input image $\mathcal{R}^{H{\times}W{\times}3}$, cropped rough mask $\mathcal{R}^{H{\times}W{\times}1}$, and cropped point heatmap $\mathcal{R}^{H{\times}W{\times}C}$) to be the input tensor $\mathcal{R}^{H{\times}W{\times}(3+1+C)}$ of the MaskRefineNet, where $C$ is the number of classes.

\section*{Appendix: Additional Analysis}

\paragraph{Effect of the input size of MaskRefineNet.}
We originally set the input size of MaskRefineNet to 256{$\times$}256. Here, we change the input size to verify its effect on the WSSIS result in table~\ref{tab:ablation_size_maskrefinenet}.
For this, we train the MaskRefineNet using the input size of 128{$\times$}128 or 384{$\times$}384.
We measure the AP result of the student network trained with the pseudo and full labels on the COCO 2017 validation set.
Consequently, the 256{$\times$}256 size yields the best AP score of 35.5\% but its performance gap with the 384{$\times$}384 size is marginal (35.5\% vs 35.4\%).

\begin{table}[t]
  \centering
  \begin{adjustbox}{max width=\linewidth}
  \begin{tabular}{c|ccc}
    \toprule
    Input Size & $AP$ & $AP_{50}$ & $AP_{75}$ \\
    \midrule
    128$\times$128 & 34.1 & 53.4 & 36.1 \\
    256$\times$256 & 35.5 & 56.0 & 37.8 \\
    384$\times$384 & 35.5 & 55.9 & 37.7 \\
    \bottomrule
  \end{tabular}
  \end{adjustbox}
  \caption{
    \textbf{Effect of the input size of MaskRefineNet}. The APs are measured on COCO 2017 validation set.
  }
  \label{tab:ablation_size_maskrefinenet}
\end{table}
\begin{table}[t]
  \centering
  \begin{adjustbox}{max width=\linewidth}
  \begin{tabular}{c|cccccc|c}
    \toprule
    Iterative & 1\% & 2\% & 5\% & 10\% & 30\% & 50\% & 100\% \\
    \midrule
               & 23.9 & 25.1 & 33.4 & 35.5 & 37.4 & 38.3 & 39.0 \\
    \checkmark & 25.6 & 26.0 & 34.5 & 35.9 & 37.6 & 38.3 & 39.0 \\
    \bottomrule
  \end{tabular}
  \end{adjustbox}
  \caption{
    \textbf{Effect of iterative training strategy.} The APs are measured on COCO 2017 validation set according to COCO subsets.
  }
  \label{tab:abs_iterative_training}
\end{table}

\begin{table}[t]
  \centering
  \small
  \begin{adjustbox}{max width=\linewidth}
  \begin{tabular}{c|c|c|c}
    \toprule
    Method & Label Types & Budget (days) $\downarrow$ & $AP$ (\%) $\uparrow$ \\
    \midrule %
    \multicolumn{4}{c}{\footnotesize 
    \textit{\textbf{Weakly-supervised Models}}} \\
    \midrule
    BBTP~\cite{(bbtp)hsu2019weakly} & $\mathcal{B}$ 100\% & 174.5 & 21.1  \\
    BBAM~\cite{(BBAM)lee2021bbam} & $\mathcal{B}$ 100\% & 174.5 & 25.7  \\
    BoxInst~\cite{(boxinst)tian2021boxinst} & $\mathcal{B}$ 100\% & 174.5 & 33.2  \\
    BoxLevelSet~\cite{(boxlevelset)li2022box} & $\mathcal{B}$ 100\% & 174.5 & 33.4  \\
    BoxTeacher~\cite{cheng2022boxteacher} & $\mathcal{B}$ 100\% & 174.5 & 35.4  \\
    Point-sup~\cite{cheng2022pointly} & $\mathcal{P}_{10}$ 100\% & 263.2 & 37.7 \\
    \midrule
    \multicolumn{4}{c}{\footnotesize \textit{\textbf{Weakly Semi-supervised Models}}} \\
    \midrule
    Ours & $\mathcal{F}$ 5\% + $\mathcal{P}$ 95\%  & 158.5 & 33.7  \\
    Ours & $\mathcal{F}$ 10\% + $\mathcal{P}$ 90\% & 196.7 & 35.8  \\
    Ours & $\mathcal{F}$ 20\% + $\mathcal{P}$ 80\% & 273.1 & 37.1  \\
    Ours & $\mathcal{F}$ 30\% + $\mathcal{P}$ 70\% & 349.5 & 38.0  \\
    Ours & $\mathcal{F}$ 50\% + $\mathcal{P}$ 50\% & 502.3 & 38.8  \\
    \midrule
    \multicolumn{4}{c}{\footnotesize \textit{\textbf{Fully Supervised Models}}} \\
    \midrule
    MRCNN~\cite{(MRCNN)he2017mask} & $\mathcal{F}$ 100\% & 884.2 & 38.8  \\
    CondInst~\cite{(condinst)tian2020conditional} & $\mathcal{F}$ 100\% & 884.2 & 39.1  \\
    SOLOv2~\cite{(solov2)wang2020solov2} & $\mathcal{F}$ 100\% & 884.2 & 39.7  \\
    \bottomrule
  \end{tabular}
  \end{adjustbox}
  \caption{
    \textbf{Additional comparisons with weakly-supervised methods in terms of labeling budget and accuracy}. We compare the methods on the COCO \textit{test-dev} under various supervisions; $\mathcal{B}$ (box label), $\mathcal{P}_{10}$ (10-points label), $\mathcal{P}$ (single-point label), $\mathcal{F}$ (full mask label). All methods use the same backbone network of ResNet-101~\cite{(resnet)he2016deep}.
  }
  \label{tab:comparison_budget_supp}
  \vspace{-2mm}
\end{table}

\paragraph{Effect of iterative training strategy.}
Some weakly-supervised methods~\cite{arun2020weakly, wang2018weakly,khoreva2017simple} utilize iterative training strategy; after training the target network, they generate pseudo labels using the target network, and then they newly train the target network using the pseudo labels.
This strategy could give additional performance improvement but demands a more complex training pipeline.
In this work, we suffer from the insufficient mask representation of the network when the amount of fully labeled data is extremely limited ($e.g.,$ COCO 1\%).
Although we can alleviate the problem with the proposed MaskRefineNet, we additionally try to adopt this strategy since we assume that the trained student network may have stronger mask representation ability than the teacher network.
For this, after training the student network, we newly generate pseudo instance masks for point labeled images.
Using both full labels and new pseudo labels, we train a new student network.
As the results in table~\ref{tab:abs_iterative_training}, the iterative training strategy yields meaningful improvements on tiny fully labeled data conditions (COCO 1\%: 23.9\%$\rightarrow$25.6\%).
However, there is no significant performance improvement for subsets above COCO 30\%.
This result demonstrates that (1) the iterative training strategy is helpful only when the amount of fully labeled data is extremely limited, (2) in more generous conditions such as COCO 30\% and 50\%, our MaskRefineNet is enough to replenish the mask representation of the network.

\noindent \textbf{Additional Comparison with weakly-supervised method}:
Point-sup~\cite{cheng2022pointly} introduced a new type of weak supervision source, multiple (10) points.
They achieved remarkable instance segmentation results with a highly reduced annotation cost.
To compare with them, we estimate the annotation time for 10-points according to the literature; they labeled 10-points in the bounding box region.
\begin{itemize}
    \item \textbf{10 Points}: $77.1 classes/img \times 1s/class+7.2inst/img \times (7s/bbox + 10 points \times 0.9s/point)=\textbf{192.3s/img}$.
\end{itemize}
In table~\ref{tab:comparison_budget_supp}, we provide the results for weakly-supervised methods and ours on COCO \textit{test-dev} in terms of accuracy and labeling budget.
Although Point-sup shows a slightly better efficiency than ours (37.7\% with a budget of 263.2 days vs. 37.1\% with a budget of 273.1 days), we argue that our training setting is more applicable for the current dataset conditions than them because they require newly annotating of 10-points.
Also, we show the possibility for more performance improvement up to 38.8\%, which is highly close to the result of the fully-supervised setting.
Furthermore, they give us a new future direction; incorporating 10-points and single-point without any mask labels.

\begin{table}[t]
  \centering
  \begin{adjustbox}{max width=\linewidth}
  \begin{tabular}{c|cccccc}
    \toprule
    Method & 5\% & 10\% & 20\% & 30\% & 40\% & 50\% \\
    \midrule
    Point DETR~\cite{(point_detr)chen2021points} & 26.2 & 30.3 & 33.3 & 34.8 & 35.4 & 35.8 \\
    Group R-CNN~\cite{(group_rcnn)zhang2022group} & 30.1 & 32.6 & 34.4 & 35.4 & 35.9 & 36.1 \\
    \midrule
    ours & 32.4 & 34.3 & 35.6 & 36.9 & 37.0 & 37.6 \\
    \bottomrule
  \end{tabular}
  \end{adjustbox}
  \caption{
    \textbf{Qualitative comparisons on COCO \textit{test-dev} object detection benchmark}.
    All methods used the ResNet-50 backbone.
  }
  \label{tab:comparision_WSSOD}
  \vspace{-2mm}
\end{table}

\noindent \textbf{Comparison with weakly semi-supervised object detection methods}: 
In our main paper, we discussed the weakly semi-supervised object detection (WSSOD) methods~\cite{(point_detr)chen2021points,(group_rcnn)zhang2022group}, which used the box labels as strong labels and the point labels as weak labels.
Since the instance segmentation covers object detection, we measure our performance on the COCO \textit{test-dev} object detection benchmark.
For this, we use the min-max points from the instance mask output as our bounding box output.
Even though our strong label is different from theirs (full mask vs. bounding box), the results in table~\ref{tab:comparision_WSSOD} show that ours can surpass the state-of-the-art WSSOD performance.
We note that all methods use the same ResNet-50~\cite{(resnet)he2016deep} backbone network and the same amount of total strong and weak labels.

\begin{figure*}[t]
    \centering
    \includegraphics[width=0.65\linewidth]{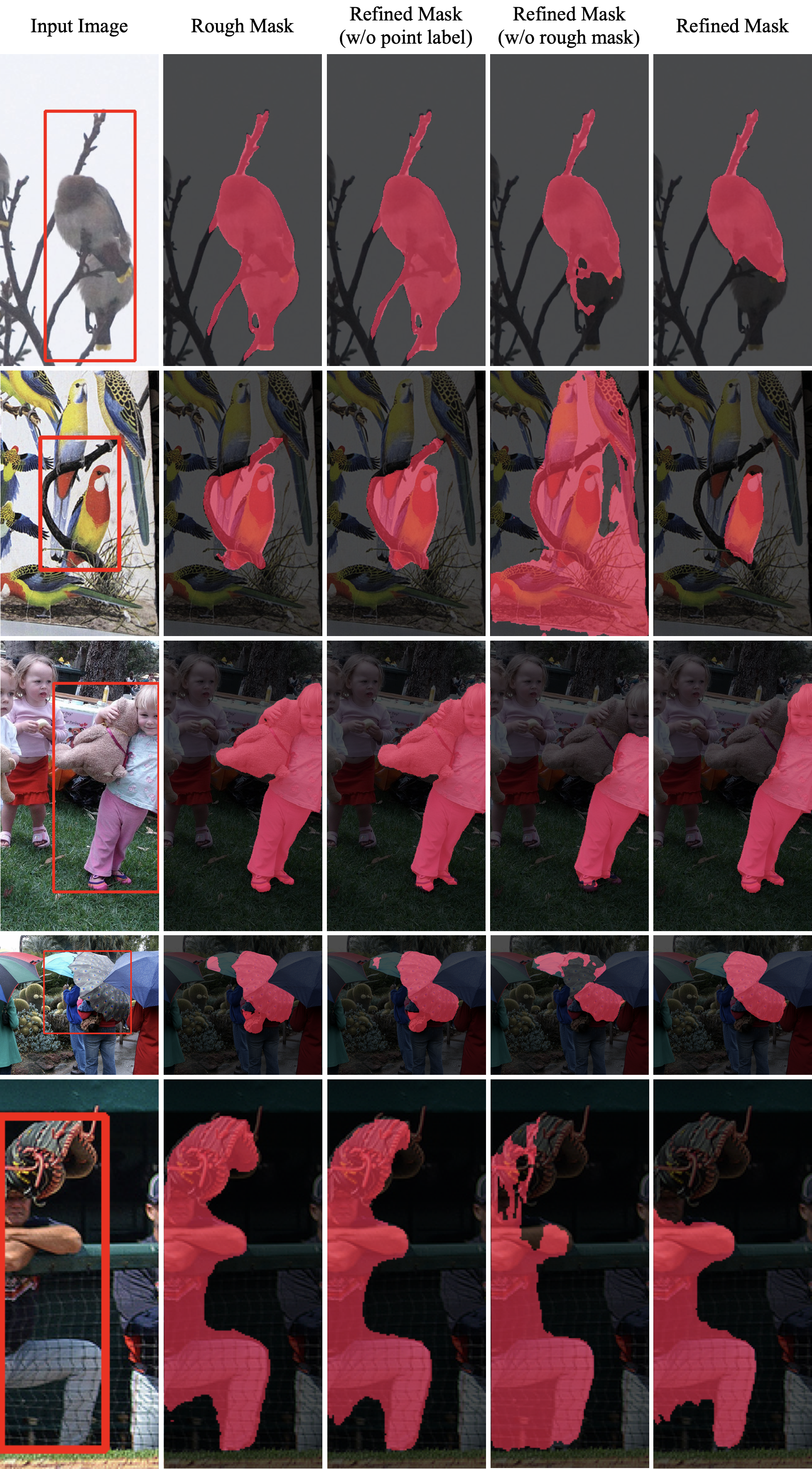}
    \caption{
        \textbf{Qualitative analysis of the effect of input sources of MaskRefineNet}. Object instances can not be distinguished when the point label is not given for MaskRefineNet (3rd col). Meanwhile, mask representations are inaccurate due to the absence of prior rough masks (4th col). Based on these low-cost priors, we can obtain sophisticated masks per object instance (5th col).
    }
    \label{fig:abs_of_maskrefinenet}
\end{figure*}

\noindent \textbf{Qualitative analysis for the effect of input sources of MaskRefineNet}: 
In Table~\ref{tab:ablation_pyramid} of our main paper, we provided the quantitative analysis of the effect of input sources of MaskRefineNet.
Here, we supplement our analysis with the qualitative results according to the input sources of the MaskRefineNet in Figure~\ref{fig:abs_of_maskrefinenet}.
When given all three informative input sources, the MaskRefineNet can produce high-quality refined masks by separating overlapping instances and removing noisy pixels.

\begin{figure*}[t]
    \centering
    \includegraphics[width=\linewidth]{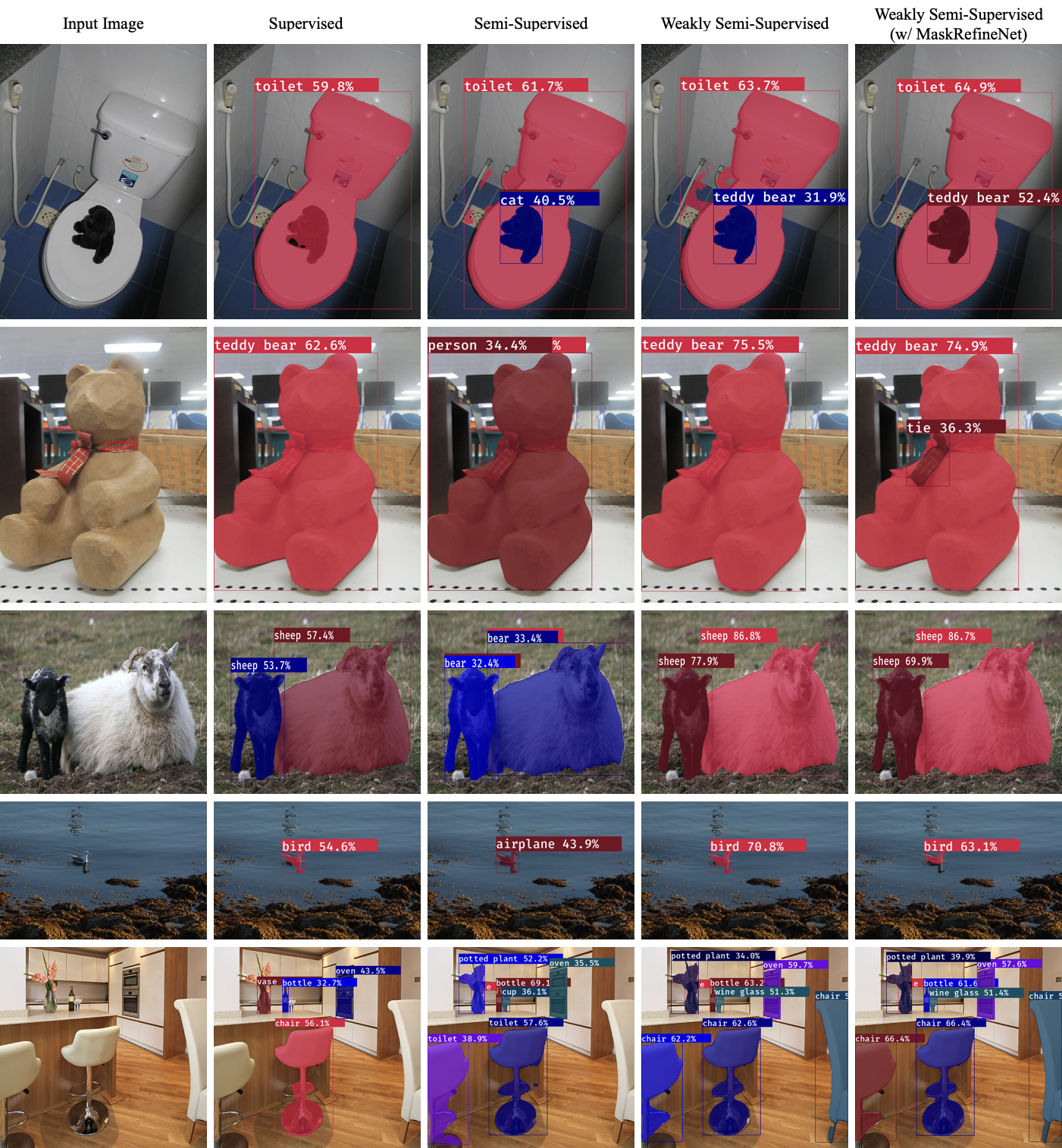}
    \caption{
        \textbf{Qualitative comparison of models trained with different types of supervision on COCO 10\% setting}. 
        The result of the semi-supervised setting can detect instance masks for all objects but is vulnerable to misclassification (e.g. cat, person, bear, airplane, toilet). Meanwhile, our point-guided model presents accurate class predictions. Our MaskRefineNet further elaborates the mask representation.
    }
    \label{fig:quality_various_supervision_sources_supp}
    \vspace{10mm}
\end{figure*}

\noindent \textbf{Qualitative comparison of baselines and our WSSIS method.} 
In Figure~\ref{fig:result_coco} of our main paper, we provided the AP evolution of two baselines and our WSSIS method according to the COCO subsets.
In Figure~\ref{fig:quality_various_supervision_sources_supp}, we provide the qualitative results of two baselines and our method under the COCO 10\% setting.
There are four types of methods: (a) training with fully labeled data only, (b) training with fully labeled data and unlabeled data, (c) training with fully labeled data and point labeled data, and (d) training with fully labeled data and point labeled data along with our point-guided MaskRefineNet.
The results demonstrate that the network trained with our method can be guided with higher-quality pseudo labels, resulting less false-positive and false-negative outputs.

\begin{figure*}[t]
    \centering
    \includegraphics[width=0.95\linewidth]{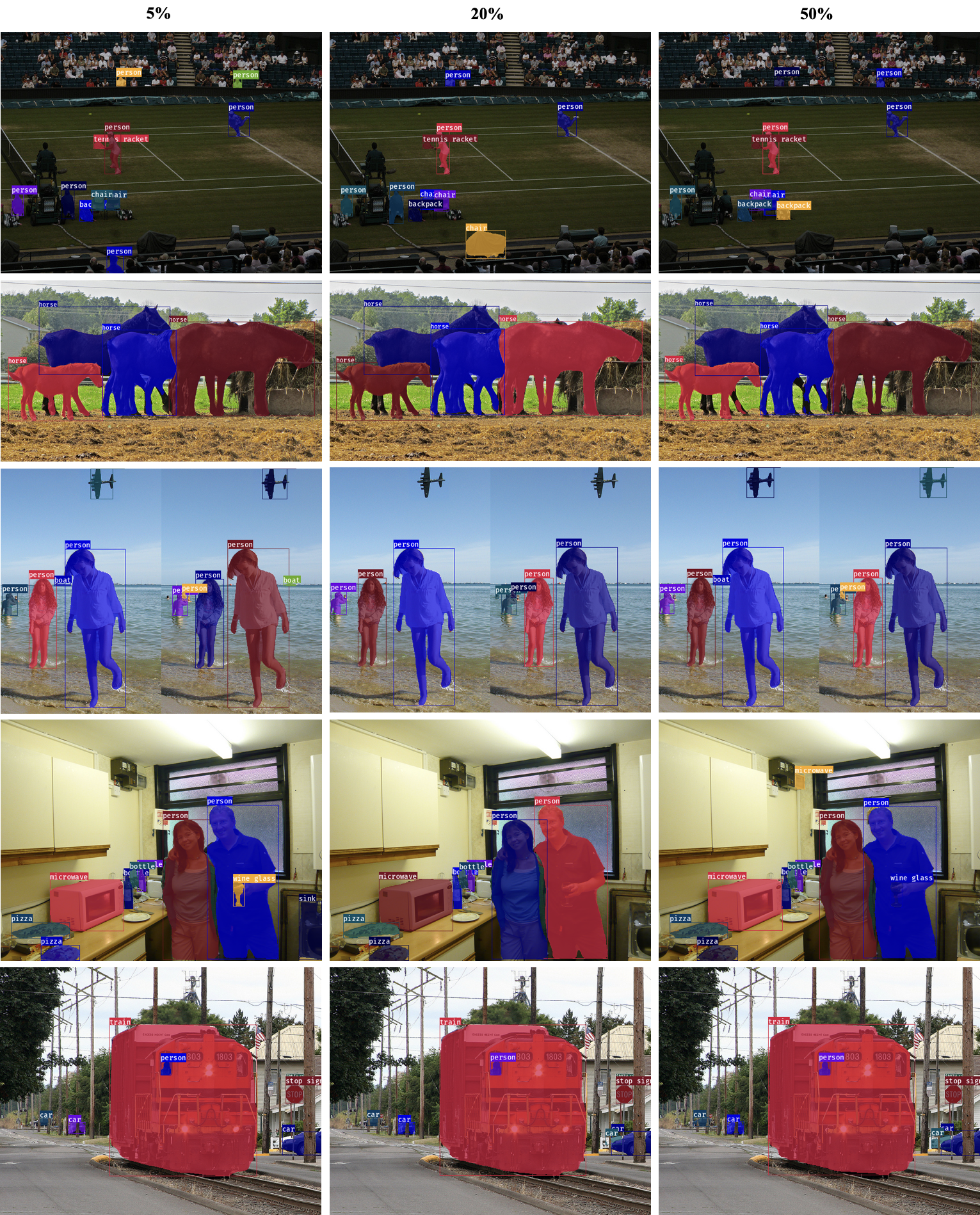}
    \caption{
        \textbf{Additional qualitative results according to the various subsets in COCO data}. Owing to our point guidance along with MaskRefineNet, leveraging only 5\% of full labeled data sufficiently localizes all the instances with elaborative mask representations.
    }
    \label{fig:abs_coco_percent_supp}
\end{figure*}

\noindent \textbf{Additional qualitative results on COCO dataset.}
In Figure~\ref{fig:abs_coco_percent_supp}, we provide additional qualitative results of ours trained with 5\%, 20\%, and 50\% COCO subsets. 

\begin{figure*}[t]
    \centering
    \includegraphics[width=\linewidth]{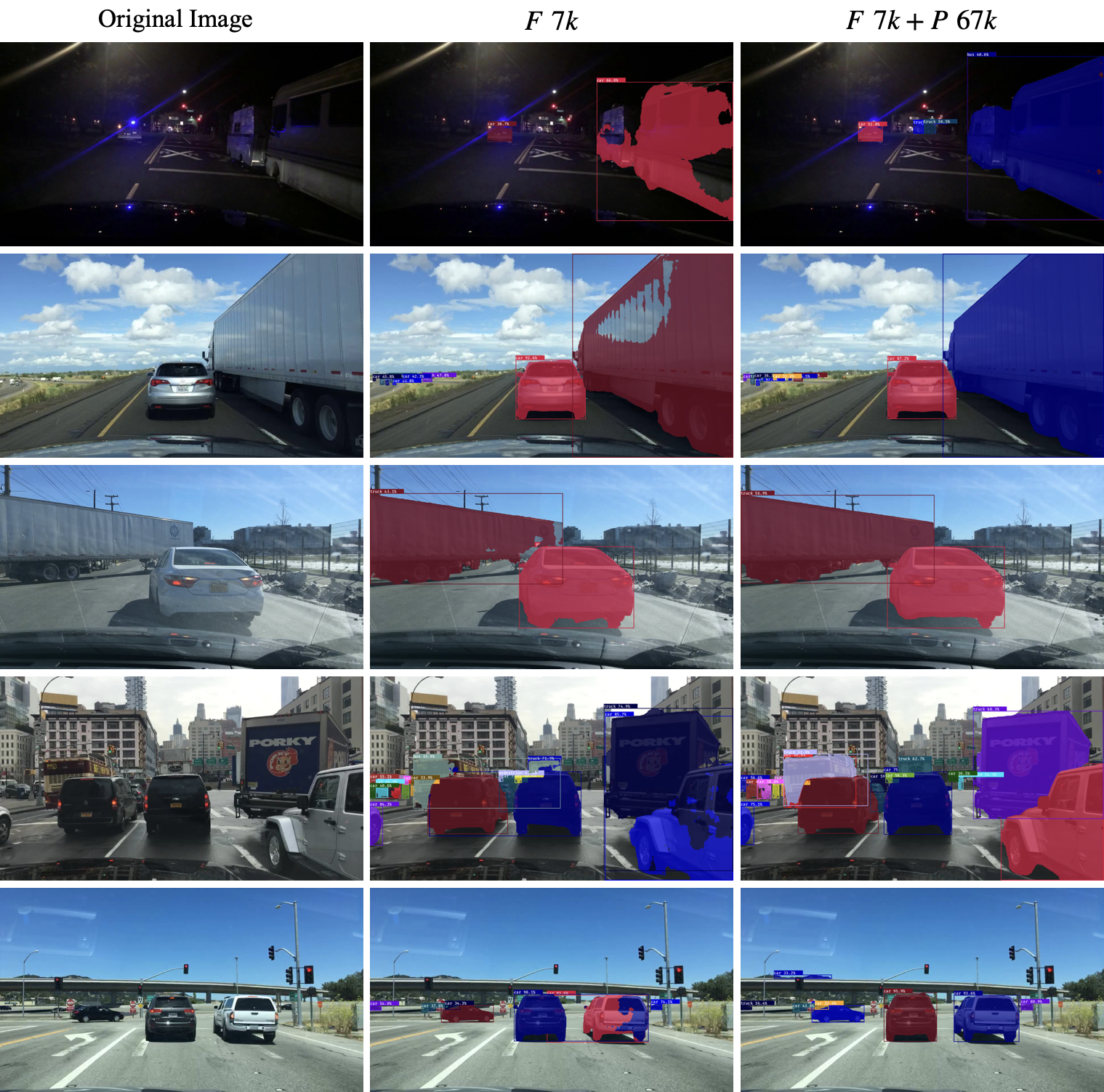}
    \caption{
        \textbf{Qualitative comparison of leveraging point labels on BDD100K}. Training with point labels clearly enriches the mask representation and removes the noise incurred by visually hard samples ($e.g.,$ dark light condition in the first row).
    }
    \label{fig:quality_bdd_supp}
\end{figure*}

\noindent \textbf{Qualitative results on BDD100K dataset.}
We qualitatively analyze the effect of leveraging point labels for the instance segmentation model using the BDD100K dataset~\cite{(bdd100k)yu2020bdd100k}.
There are two types of networks: the first is the network trained with only 7K fully labeled data, and the second is the network trained with 7K fully labeled data and 67K point labeled data.
As shown in Figure~\ref{fig:quality_bdd_supp}, due to our effective leveraging of the point labels, the second network is much more robust to large and small instances and occluded instances.

\nocite{(wssod)yan2017weakly}
\nocite{lee2013pseudo}
\nocite{shen2019cyclic}
\nocite{khoreva2017simple}
\nocite{hu2018learning}
\nocite{(prm)zhou2018weakly}
\nocite{(liid)liu2020leveraging}
\nocite{arun2020weakly}
\nocite{lee2019ficklenet}
\nocite{lee2021anti}
\nocite{ouali2020semi}
\nocite{jeong2019consistency}
\nocite{he2021re}
\nocite{(boxlevelset)li2022box}
\nocite{cheng2022boxteacher}
\nocite{(DRS)kim2021discriminative}
\nocite{(SSUL)cha2021ssul}

{\small
\bibliographystyle{ieee_fullname}
\bibliography{ms}
}

\end{document}